\documentclass[10pt, conference, compsocconf]{IEEEtran}
\ifCLASSINFOpdf
\else
\fi

\usepackage{microtype}      % microtypography
\usepackage{times}
\usepackage{hyperref}
\usepackage{epsfig}
\usepackage{graphicx}
\usepackage{url}
\usepackage{color}
\usepackage{mathrsfs}  
\usepackage{amssymb,amsmath,bm}
\usepackage{multicap,graphicx}
\usepackage{algorithm}  
\usepackage{algorithmic}
\usepackage{graphicx}
\usepackage{subcaption}
\usepackage{wrapfig}
\usepackage{picinpar}
\usepackage{cutwin}
\usepackage{stmaryrd}

\begin{document}
%
% paper title
% can use linebreaks \\ within to get better formatting as desired
\title{Modified Distribution Alignment for Domain Adaptation  with Pre-trained Inception ResNet}

% author names and affiliations
% use a multiple column layout for up to two different
% affiliations

\author{\IEEEauthorblockN{Authors Name/s per 1st Affiliation (Author)}
\IEEEauthorblockA{line 1 (of Affiliation): dept. name of organization\\
line 2: name of organization, acronyms acceptable\\
line 3: City, Country\\
line 4: Email: name@xyz.com}
}

\author{\IEEEauthorblockN{Youshan Zhang}
\IEEEauthorblockA{Computer Science and Engineering Department\\
Lehigh University\\
Bethlehem, PA\\
yoz217@lehigh.edu}
\and
\IEEEauthorblockN{Brian D. Davison}
\IEEEauthorblockA{Computer Science and Engineering Department\\
Lehigh University\\
Bethlehem, PA\\
davison@cse.lehigh.edu}
}

% conference papers do not typically use \thanks and this command
% is locked out in conference mode. If really needed, such as for
% the acknowledgment of grants, issue a \IEEEoverridecommandlockouts
% after \documentclass

% for over three affiliations, or if they all won't fit within the width
% of the page, use this alternative format:
% 

% make the title area
\maketitle

\begin{abstract}
Deep neural networks have been widely used in computer vision. There are several well trained deep neural networks for the ImageNet classification challenge, which has played a significant role in image recognition. However, little work has explored pre-trained neural networks for image recognition in domain adaption. In this paper, we are the first to extract better-represented features from a pre-trained Inception ResNet model for domain adaptation. We then present a modified distribution alignment method for classification using the extracted features. We test our model using three benchmark datasets (Office+Caltech-10, Office-31 and Office-Home). Extensive experiments demonstrate significant improvements (4.8\%, 5.5\%, and 10\%) in classification accuracy over the state-of-the-art. 

\end{abstract}


% For peer review papers, you can put extra information on the cover
% page as needed:
% \ifCLASSOPTIONpeerreview
% \begin{center} \bfseries EDICS Category: 3-BBND \end{center}
% \fi
%
% For peerreview papers, this IEEEtran command inserts a page break and
% creates the second title. It will be ignored for other modes.
\IEEEpeerreviewmaketitle

\section{Introduction}
 With the rapid development of social media and content sharing applications, data grows much faster than we can make sense of it. There is  great demand for automatic classification and analysis for text, images, and other multimedia data \cite{chen2018zero}. However, it is time-consuming and expensive to acquire enough labeled data to train models.  Therefore, it is valuable to learn a model for a new target domain from a different domain with abundant labeled samples.  Mechanisms for learning feature representations of a continuous intermediate space from one domain to another has been widely used in many fields such as machine learning \cite{ben2007analysis}, language processing \cite{blitzer2007biographies}, and computer vision \cite{gopalan2011domain}. There are several techniques to address this problem; a prominent one is domain adaption \cite{gong2012geodesic,long2013transfer,tzeng2014deep}. There have been efforts for both semi-supervised \cite{pan2011domain,ben2010theory,bergamo2010exploiting} and unsupervised \cite{blitzer2008learning,mansour2009domain,ben2010impossibility} domain adaptation. In the first case, the target domain contains a small amount of labeled data; for the latter case, the target domain is entirely unlabeled. Usually the labeled target data alone is insufficient to construct a good classifier. Thus, how to effectively leverage sufficient label source data to facilitate unlabeled target data is key to domain adaptation.

However, a critical challenge remains: to find and identify useful features that span the representations of two domains. The quality of such features will directly affect classification accuracy. We cannot expect to train a high-quality classifier if the learned features are poor. Therefore, it is essential to find a proper way to represent the source and target data.

One useful working model for feature representation is based on manifold learning, which learns the intermediate features between the source and the target domain via a Grassmannian manifold.  Gopalan et al.\ \cite{gopalan2011domain} proposed a sampling geodesic flow (SGF) method to learn the intermediate features between the source and the target domain via the geodesic (shortest path) on Grassmannian manifold. However, Gong et al.\ \cite{gong2012geodesic} have noted that it is difficult to choose an optimal sampling strategy. Moreover, SGF has high time complexity making sampling slow when many points are needed. Gong et al.~\cite{gong2012geodesic} proposed a geodesic flow kernel (GFK) model to overcome the limitations of unknown sampling size in SGF. 
They integrated all samples along the ``geodesic", which is calculated from Gopalan et al.\ \cite{gopalan2011domain}. We show that the ``geodesic" is not the true geodesic. Several works have addressed the alignment of marginal distribution and conditional distribution of data in domain adaption. Wang and Mahadevan aligned the source and target domain by preserving the ‘neighborhood structure’ of the data   points \cite{wang2009manifold}.  Wang et al.\ proposed a manifold embedding distribution alignment method (based on work of Gong et al.~\cite{gong2012geodesic}) to align both the degenerate feature transformation and the unevaluated distributions of both domains \cite{wang2018visual}. However, none of these models explore the quality of the learned features.
%\cite{gong2012geodesic,wang2018visual}

Deep learning models are also widely applied to domain adaptation
%features from source to target domain 
\cite{tzeng2014deep,sun2016deep,tzeng2017adversarial,long2017deep,jiang2017integration,chen2018joint,zhang2018collaborative,chen2018joint}. Stacked Denoising Autoencoders is one of the first deep models for domain adaptation, and aims to find the common features between the source and target domain via denoising autoencoders \cite{vincent2008extracting}. The deep neural network for domain adaptation can be majorly classified in four types: discrepancy-based methods, adversarial discriminative models, adversarial generative models, and data reconstruction-based models. One of the first discrepancy-based methods is Deep Domain Confusion (DDC), which considers the discrepancy in different layers and the network is fine-tuned based on maximum mean discrepancy (MMD) \cite{tzeng2014deep}. Later Long et al.~\cite{long2015learning} proposed a Deep Adaptation Network (DAN) that considered the sum of MMD from several layers with several 
%different 
kernels of MMD functions. The Domain adaptive neural network also embedded MMD as a regularization \cite{ghifary2015domain}. Adversarial discriminative based models aim to define a domain confusion objective to identify the domains via a domain discriminator. The Domain-Adversarial Neural Networks (DANN) consider a minimax loss to integrate a gradient reversal layer to promote the discrimination of source and target domain \cite{ganin2016domain}.
The Adversarial Discriminative Domain Adaptation (ADDA) uses an inverted label GAN loss to split the source and target domain, and features can be learned separately \cite{tzeng2017adversarial}. The adversarial generative models combine the discriminative model with generative components based on Generative Adversarial Networks (GANs) \cite{goodfellow2014generative}.  The Coupled Generative Adversarial Networks \cite{liu2016coupled} consists of a series of GANs, and each of them can represent one of the domains. Data reconstruction-based methods jointly learn source label predictions and unsupervised target data reconstruction \cite{bousmalis2016domain}. 
% Domain Separation Networks (DSN) \cite{bousmalis2016domain} introduces the notion of a private subspace for each domain, which captures domain-specific properties, such as background and low-level image statistics. A shared subspace, enforced through the use of autoencoders and explicit loss functions, captures common features between the domains. 

However, training of deep neural networks consume time and require much effort to tune the parameters. We are inspired by Zhang et al.~\cite{zhang2018automated}, which extracted features from the well-trained Alexnet, and then trained an SVM using the deep features to facilitate improvements in classification accuracy. Also, other work indicated that the features extracted from the activation layers of a well-trained deep neural network could be re-used for different tasks even when the new tasks are different from the original tasks used to train the model \cite{donahue2014decaf}.

In this paper, we first extract features from a well-trained Inception ResNet-v2 (IR) model; we then classify these features based on a modified distribution alignment. Our contributions are three-fold:
\begin{enumerate}
    \item We create three datasets for domain adaptation based on better extracted features, which can be of significant value in future research for the community. 
    \item We show the shortcomings of the original manifold embedded distribution  alignment method, and propose a modified distribution alignment for classification, which enhances the accuracy for  classification.
    \item We test these improvements using three benchmark datasets. Extensive experiments demonstrate significant improvements (4.8\%, 5.5\%, and 10\%) in classification accuracy over the state-of-the-art.  
\end{enumerate} 
\section{Problem Statement} \label{sec:class}

To avoid the complex and time-consuming process of hand-tuning  parameters for training a deep neural network, we present the extraction of features from a well-trained deep neural network, so that we are able to learn a better feature representation of source and target domain data. Also, we want to further align the distribution from both source and target domain.

Given training data (source domain): $X_S$, with its labels $Y_S=\{y_{i}\}_{i=1}^{N_1} \in \{1, 2, 3, \cdots, C\}$, denoting the $C$ categories, and the test data (target domain): $X_T$ with its labels $Y_T=\{y_{i}\}_{i=1}^{N_2} \in \{1, 2, 3, \cdots, C\}$ and $N_2 \leq N_1$,  that implies that we might not have all labels for testing data. If $N_2=N_1$, which means we have sufficient labels for $X_T$, we aim to get a higher predictive accuracy. If $N_2< N_1$, we not only want to get a high enough predictive accuracy, but also to predict the labels for the unlabeled data. We have two concerns: 1) how to generate better source $X_S$ and target $X_T$ features for the image recognition problem; 2) how to improve prediction accuracy using the features of step 1. 

\section{Method}

\begin{figure}[t]
\centering
\includegraphics[scale=0.35]{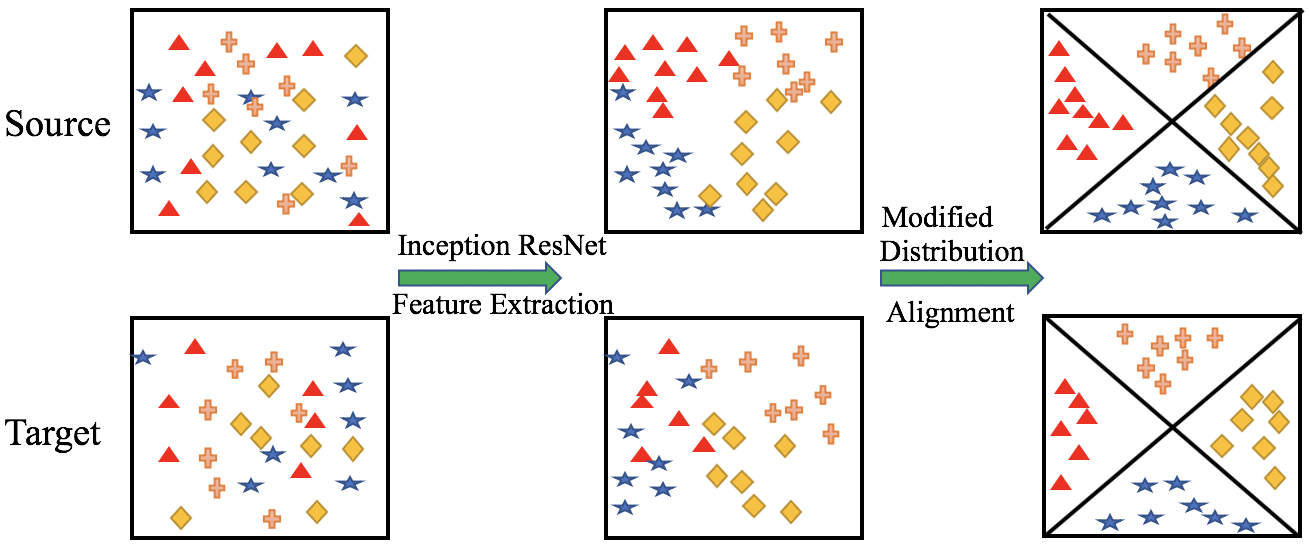}
\caption{The scheme of MDAIR model. 1) We first extract the feature from the last fully connected layer in Inception-ResNet-v2 model. The learned features are slightly more aligned than the raw features; 2)  We then align the distribution of learned features.}
\label{fig:DAIR}
\end{figure}

\subsection{Feature Extraction}
Feature extraction is a relatively easy and fast way to take advantage of deep learning without investing time and much effort into training a full neural network. Feature extraction will be especially useful if we do not use GPUs since it only requires a single pass over the input images. Kornblith et al.\  indicated that ResNets are often the best feature extractors, independently of their ImageNet accuracies \cite{kornblith2018better}. In this paper, we use Inception-ResNet-v2 as the pre-trained model from which to extract features. Inception-ResNet-v2 is a powerful convolution neural network, which is trained on more than one million images from the ImageNet datasets. This network consists of 164 layers (the largest number of convolutional and fully connected layers from the input layer to the output layer). IR model can predict 1000 categories of images, such as cup, smart phone, backpack, and many animals. Therefore, IR model has learned rich feature representations with a wide range of images. The image input size of IR model is 299-by-299-by-3. Please refer to \cite{szegedy2017inception} for details of Inception-ResNet-v2  model.

\begin{figure}[h]
\centering
\includegraphics[scale=0.44]{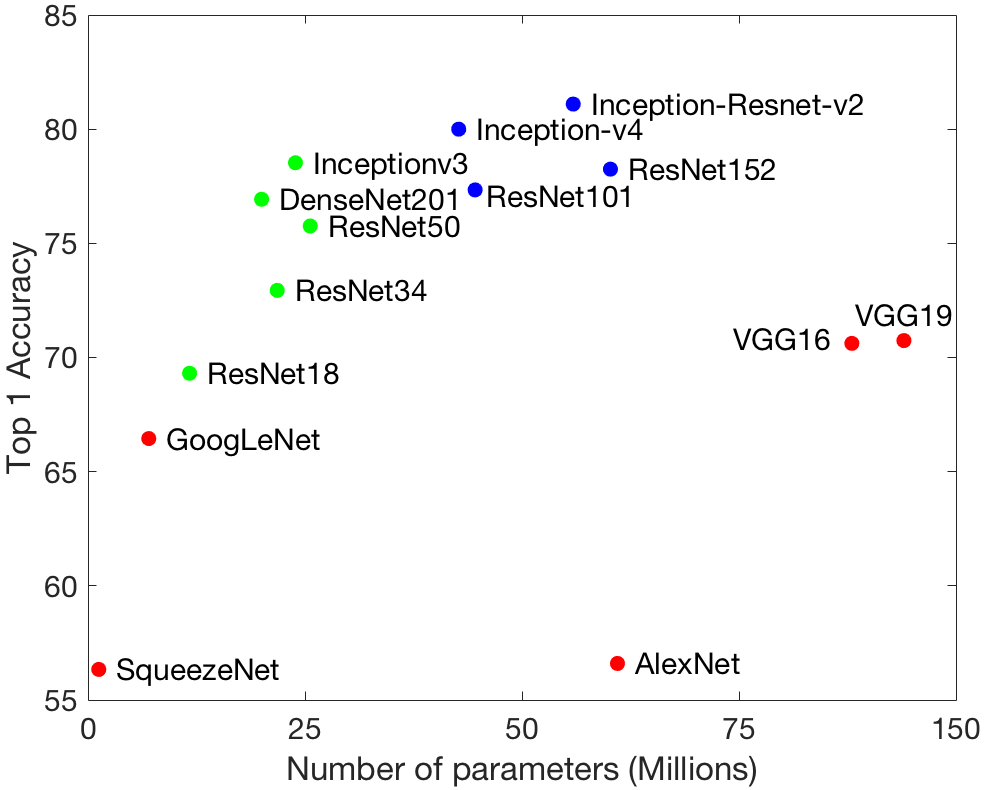}
\caption{The top 1 accuracy and number of parameters of different pre-trained deep neural networks. }
\label{fig:top-1}
\end{figure}

As shown in Fig.~\ref{fig:top-1}, we compare the number of parameters and top-1 accuracy of several well-trained deep neural networks (SqueezeNet \cite{iandola2016squeezenet} , AlexNet \cite{krizhevsky2012imagenet}, VGG16  \cite{simonyan2014very}, VGG19 \cite{simonyan2014very}, GoogLeNet  \cite{szegedy2015going}, ResNet18 \cite{he2016deep}, ResNet50, ResNet101, ResNet152 \cite{he2016deep}, DenseNet201 \cite{huang2017densely}, Inceptionv3 \cite{szegedy2016rethinking}, Inception-Resent-V2   \cite{szegedy2017inception}).
There are two essential reasons why we choose the IR model as the deep neural network to extract features. First, the top-1 accuracy of Inception-ResNet-v2 model is higher than other models. Secondly, the IR model uses fewer parameters compared with several lower accuracy networks (e.g. VGG-16).
 
\begin{figure*}[h]
\centering
\begin{subfigure}{0.35\textwidth}
\includegraphics[width=\linewidth]{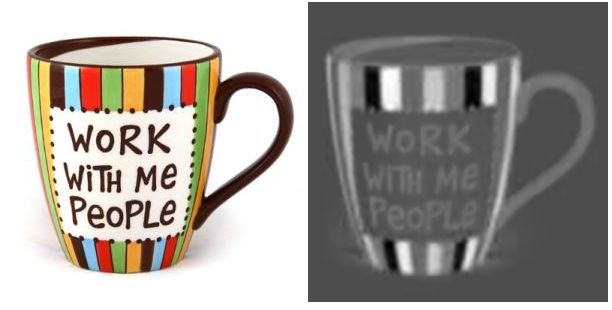}
\caption{Original image and strongest  channel} \label{fig:ima}
\end{subfigure}
\begin{subfigure}{0.35\textwidth}
\includegraphics[width=\linewidth]{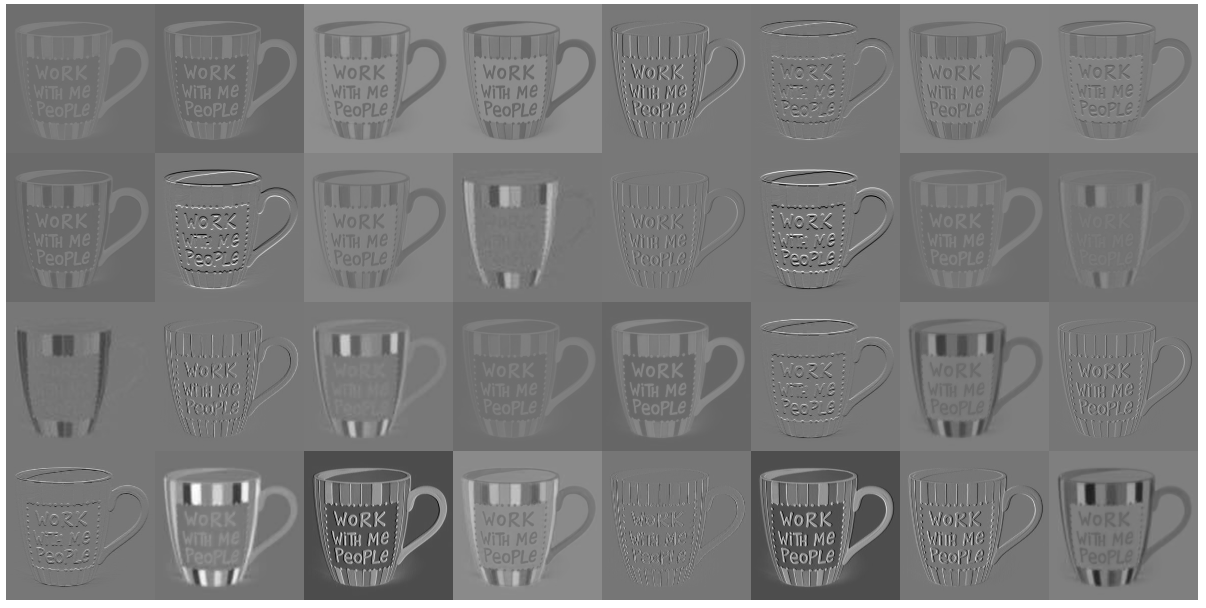}
\caption{The activation of first convolutional layer } \label{fig:imb}
\end{subfigure}
\begin{subfigure}{0.22\textwidth}
\includegraphics[width=\linewidth]{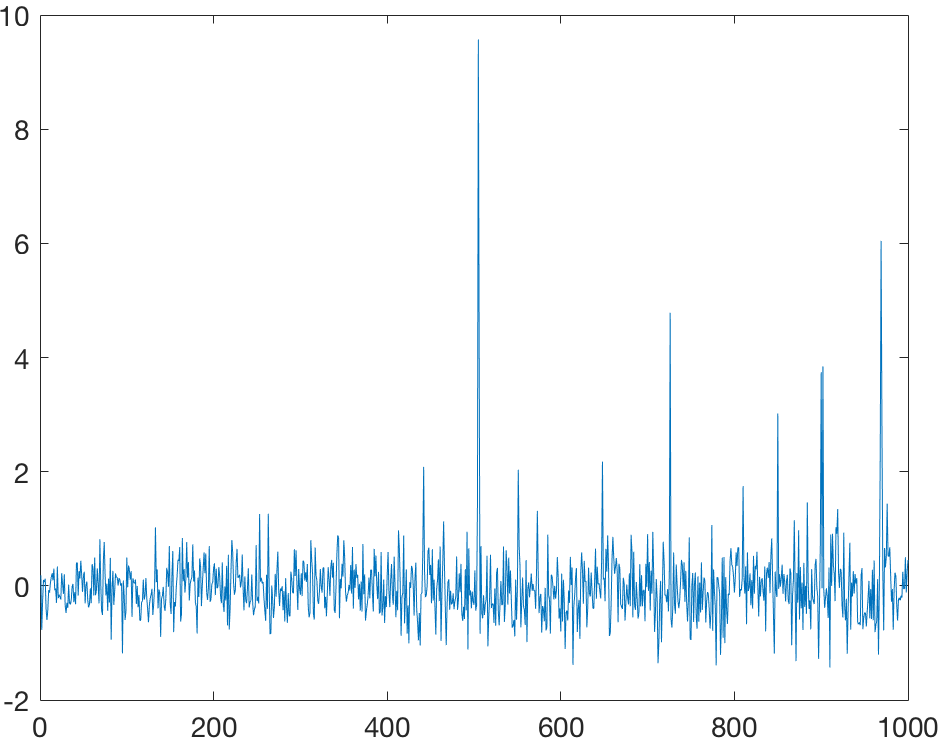}
\caption{Final extracted feature} \label{fig:imc}
\end{subfigure}
\caption{Original image and extracted features of first convolutional layer in Inception-ResNet-V2 model.  } \label{fig:mug}
\end{figure*}

We assume that extracted features from the IR model contain more detailed information than other features, which will enable a classifier to achieve higher accuracy. We then compare extracted IR features with three commonly used sets of features (SURF, Resnet-50, and DeCAF), which is shown in Sec.~\ref{sec:fea_c}.  In addition, the extracted features from different layers will have different effects on final recognition results, which is also shown in Sec.~\ref{sec:fea_c}.  Fig.~\ref{fig:mug} is an example of extracting features using the well-trained Inception-ResNet-v2 model. The left of Fig.~\ref{fig:ima} is the input image, Fig.~\ref{fig:imb} is the extracted features from the first conventional layer in the IR model; the right of Fig.~\ref{fig:ima} is strongest channel feature in Fig.~\ref{fig:imb}; and Fig.~\ref{fig:imc} is extracted feature from last fully connected layer. Alg.~\ref{alg:Alg.1} describes the procedures of extracting features from the pre-trained IR model.                                                                                      

\begin{algorithm}[h]  
\small
\caption{Extracting features from IR model} \label{alg:Alg.1}  
\begin{algorithmic}[1]
\REQUIRE{Raw images and pre-trained Inception-ResNet-v2 model}
\ENSURE{Extracted features from IR model}
\STATE 
Prepare the images (rescale the size of images to be $299\times 299 \times 3$)
\STATE 
Select one layer to extract features
% \REPEAT   
\STATE 
Apply the raw features of a datapoint as input, and use activation functions to extract the feature using IR model in the selected layer \\
\end{algorithmic}  
\end{algorithm}

\subsection{Distribution Alignment}
To train a robust classifier for features, which were extracted in the previous section, we perform dynamic distribution alignment to quantitatively account for the relative importance of marginal and conditional distribution to address the challenge of unevaluated distribution alignment.

Manifold Embedded Distribution Alignment (MEDA) is proposed by Wang et al.~\cite{wang2018visual} to align learned features from manifold learning. 
% Here, we introduce the main idea of MEDA.
It has three fundamental steps: 1) learn features from the manifold based on Gong et al.~\cite{gong2012geodesic}; 2) use dynamic distribution alignment to estimate the marginal and conditional distributions of data; and, 3) update the classified labels based on estimated parameters.
% construct a new classifier from previous steps. 
Please refer to Wang et al.~\cite{wang2018visual} for more details. The classifier ($fr$) is defined as:

\begin{equation}
\begin{aligned}\label{eq:meda}
fr= \mathop{\arg\min}_{fr \in  \mathcal{H}_{k}} & \sum_{i=1}^{N_1} \mathop{l} (fr( g(X_{S_{i}})),Y_{S_i})+\eta ||fr||_{K}^{2}\\ & + \lambda \overline{D_{fr}} (X_{S},X_{T} ) + \rho R_{fr}(X_{s},X_{T} )
\end{aligned}
\end{equation}
where $\mathcal{H}_{k}$ represents kernel Hilbert space; $l(\cdot, \cdot)$ is the loss function; $g(\cdot)$ is a feature learning function in Grassmannian manifold \cite{gong2012geodesic}; $X_{S}$ is the learned features from IR model, $||fr||_K^2$ is the squared norm of $fr$;  $\overline{D_{fr}}(\cdot, \cdot)$ represents the dynamic distribution alignment; $R_{fr}(\cdot, \cdot)$ is a Laplacian regularization; $\eta$, $  \lambda$, and $\rho$ are regularization parameters. Here, the term $\mathop{\arg\min}_{fr \in  \mathcal{H}_{k}}  \sum_{i=1}^{N_1} \mathop{l} (fr( g(X_{S_{i}})),Y_{S_i})+\eta ||fr||_{K}^{2}$ is the structure risk minimization (SRM). We can only employ the SRM on $X_S$, since there are few labels (perhaps no labels) for $X_T$.
By training the classifier from Eq.~\ref{eq:meda}, we can predict labels of test data.  

\subsection{Weaknesses of MEDA}
The first step of the MEDA method is learning the kernel mapping $g(\cdot)$ from Grassmannian manifold based on GFK model. However, the calculation of ``geodesic" in GFK model is originally from SGF method, which is a unevaluated geodesic \cite{gopalan2011domain}. Then GFk considered all samples points on ``geodesic" for constructing a kernel function. It is a ``kernel trick"; but it cannot maintain the true information from a manifold since geodesic is not correctly estimated. We design two experiments to show the defects of GFK.

\begin{figure}[h]
\centering
\includegraphics[scale=0.45]{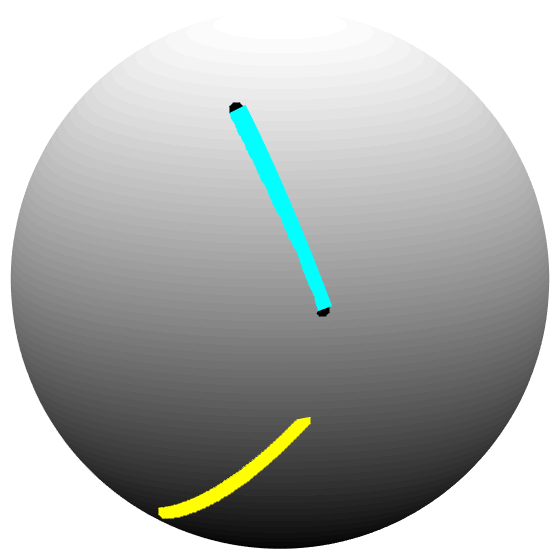}
\caption{The comparison of SGF samples and ground truth. Two black points are the given points; the cyan curve highlights the true geodesic points; the yellow curve is the sampling results of SGF. Sampled points are away from the true geodesic in SGF model. }
\label{fig:sphere}
\end{figure}

Given two points $P_1$ and $P_2$ on the sphere, we want to recover all other points between them. As shown in Figure~\ref{fig:sphere}, sampled points of the SGF method (yellow curve is not able to recover the true points on a geodesic (cyan curve). Therefore, the GFK model will lose feature information if it integrates all pseudo samples from wrong geodesic (yellow curve), which is calculated using the SGF method.   

\begin{figure}[h]
\centering
\begin{subfigure}{0.41\textwidth}
\includegraphics[width=\linewidth]{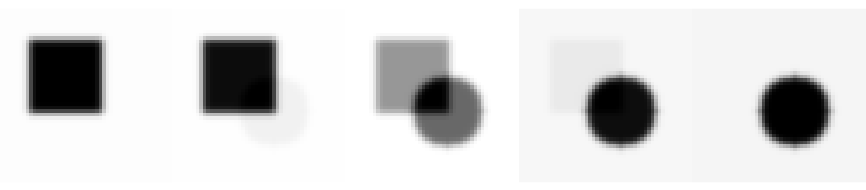}
\caption{The progress of true samples } \label{fig:true}
\end{subfigure}
\begin{subfigure}{0.41\textwidth}
\includegraphics[width=\linewidth]{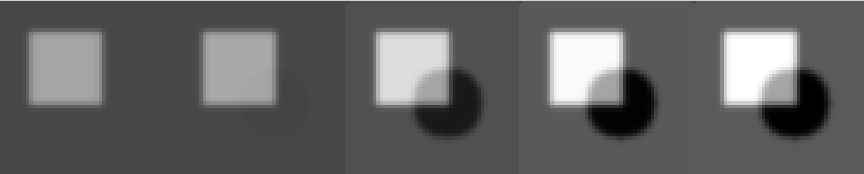}
\caption{The progress of SGF samples} \label{fig:SGF}
\end{subfigure}
\caption{The comparison of sampling results between the two images (square and circle) with $t=0,$ $0.05, 0.5, 0.95, 1$. Obviously, the SGF model does not generate a correct sample in (b). For reference, the source image is the far left at $t=0$ and the target image is  far away at $t=1$ in Fig.~\ref{fig:true}. } \label{fig:square}
\end{figure}

We design another experiment to show shape deformation using SGF model. As shown in  Fig.~\ref{fig:square}, the source image is a square (the leftmost of Fig.~\ref{fig:true}), and the target image is a circle (the rightmost of Fig.~\ref{fig:true}).  The progress of sampled images of the SGF model are shown in Fig.~\ref{fig:SGF}.  To evaluate the quality of samples, there are two criteria. The sample should be similar to the source image when $t=0$, and the sample should be similar to the target image when $t=1$.  However, the sampled images of SGF model are far from the source and target images when $t=0.05$ and $t=0.95$, respectively.

There are two issues in the sampled images of the SGF method: first, its background is dark; this is caused by the $\text{Log}$ map not being correctly calculated in Gopalan et al.~\cite{gopalan2011domain} (there are some negations of the estimated velocity $v$ between the source image and target image). The second is that the shape is never unified, and this is caused by the $\text{Exp}$ does not approach the target at $t=1$ \cite{gopalan2011domain}.

\begin{figure*}[t]
\centering
\begin{subfigure}{0.3\textwidth}
\includegraphics[width=\linewidth]{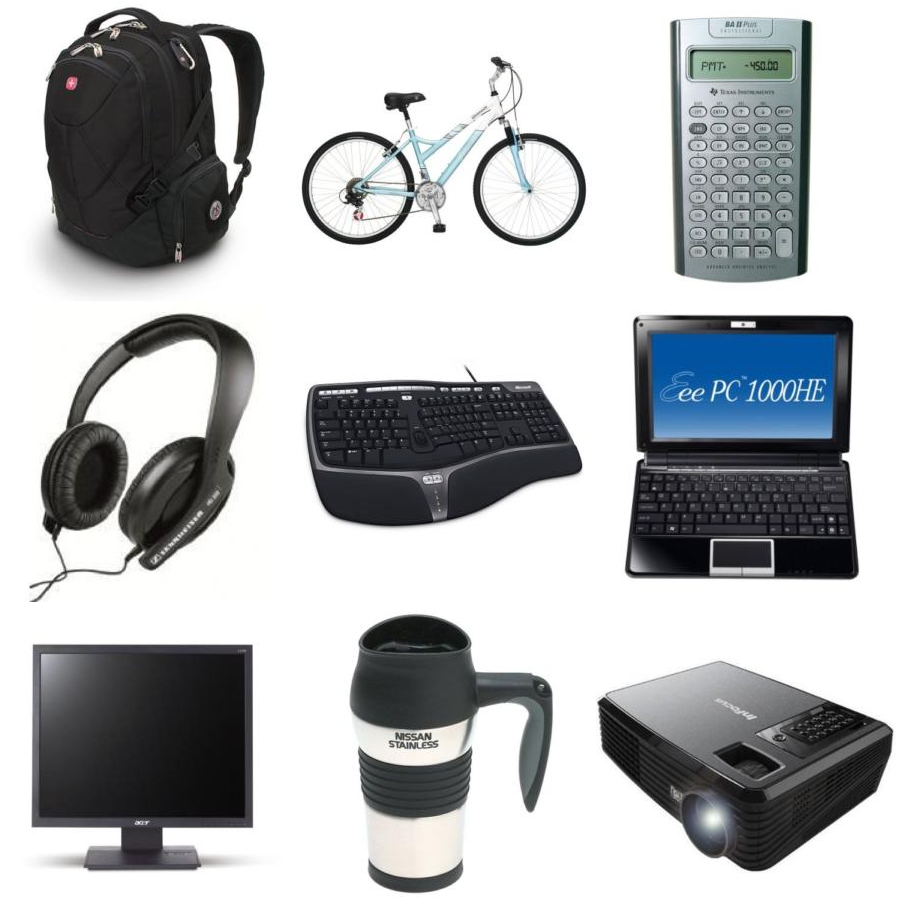}
\caption{Office+Caltech-10}
\end{subfigure}
\begin{subfigure}{0.3\textwidth}
\includegraphics[width=\linewidth]{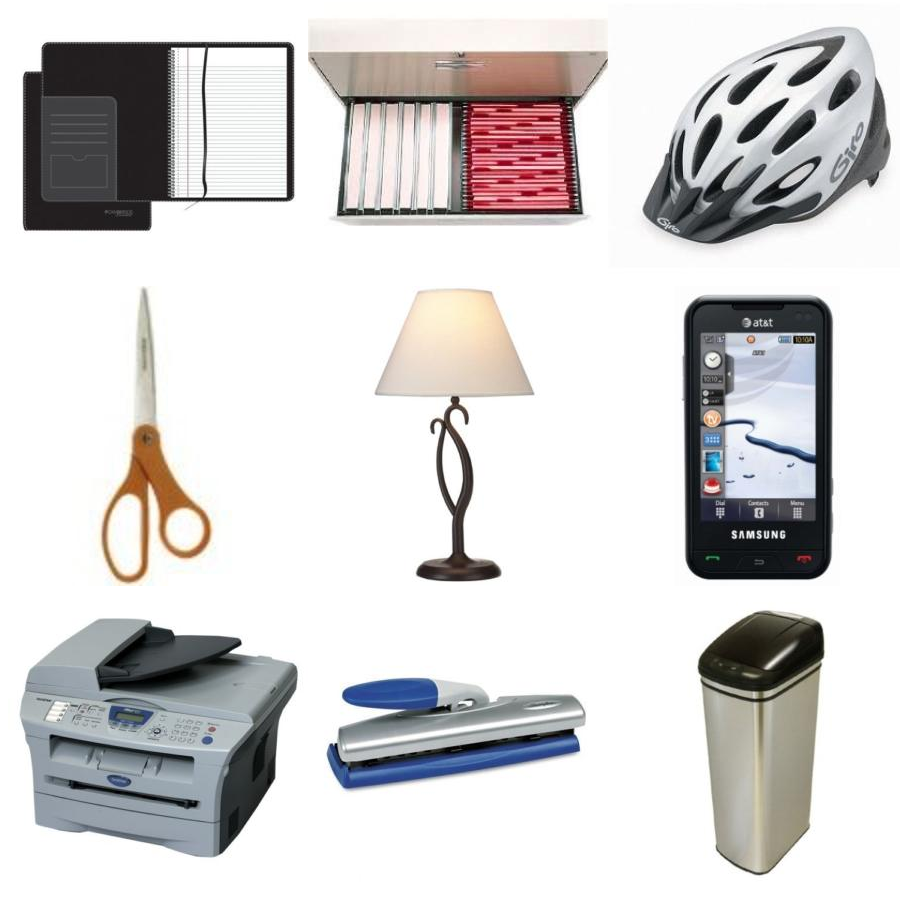}
\caption{Office-31}
\end{subfigure}
\begin{subfigure}{0.3\textwidth}
\includegraphics[width=\linewidth]{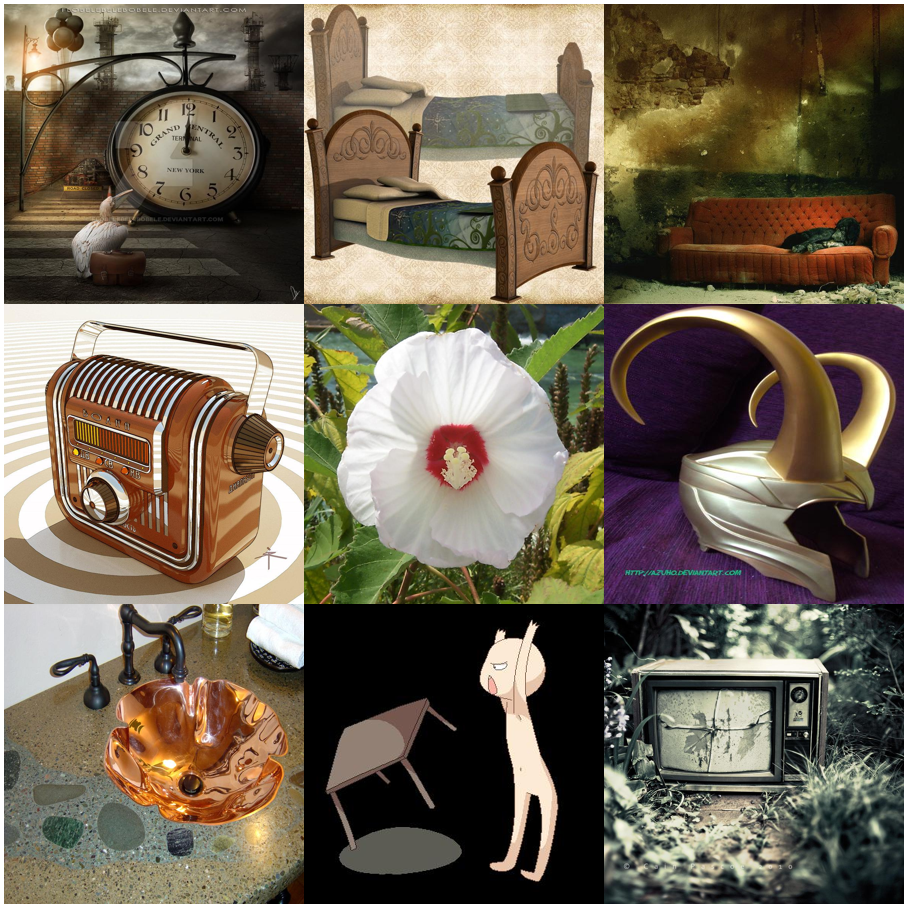}
\caption{Office-Home}
\end{subfigure}
\caption{Some example images from three benchmark datasets. (a) is from the DSLR domain in Office+Caltech-10 dataset; (b) is from the Amazon domain in Office-31 dataset, and (c) is from Art domain in Office-Home dataset.} \label{fig:images}
\end{figure*}

The second shortcoming of GFK is that the dimensionality is difficult to determine. The first step of GFK is to project the original source and target data into a subspace since the number of instances in the original space is not the same ($N_2 \leq N_1$). The reduced dimensionality will lead to information loss of original data.

\subsection{Modified Distribution Alignment}
To resolve the issues mentioned above, we use the original features instead of features from the GFK model. These are two essential reasons: 1) we want to maintain the information of original features, and we want to avoid the undetermined dimensionality in the GFK model; 2) the extracted IR features contain enough detailed information for the classification problem\footnote{Source code is available at: \url{https://github.com/heaventian93/MDAIR}.}.
Therefore, we have the following objective function:
\begin{equation}
\begin{aligned}\label{eq:MDA}
f= \mathop{\arg\min}_{f \in  \mathcal{H}_{k}} & \sum_{i=1}^{N_1} \mathop{l} (f( X_{S_{i}}),Y_{S_i})+\eta ||f||_{K}^{2}\\ & + \lambda \overline{D_{fr}} (X_{S},X_{T} ) + \rho R_{f}(X_{s},X_{T} )
\end{aligned}
\end{equation}

We only need to replace the manifold learning feature $Z$ in line 1 of Alg.1 in Wang et al.~\cite{wang2018visual} with our extracted IR features to get the modified distribution alignment model.

% We get the following algorithm:

% \begin{algorithm}[h]  
% \small
% \caption{Classification with modified distribution alignment using IR feauture} \label{alg:Alg.2}  
% \begin{algorithmic}[1]
% \REQUIRE{Data: $X_S$, $X_T$. $Y_S$, $Y_T$, and parameters $\lambda$, $\eta$, $\rho$ and $p$. }
% \ENSURE{Classification accuracy}
% \STATE 
% Replace the feature $Z$ in the line 1 of of Alg.1 in \cite{wang2018visual}
% \STATE 
% Calculate the accuracy use the $f$. \\
% \end{algorithmic}  
% \end{algorithm} 

\section{Results}

\subsection{Description of Datasets}
In this experiment, we show how our MDAIR method can enhance image recognition accuracy. We test our model using three public image datasets:  Office+Caltech-10 (we combine Office-10 and Caltech-10 as one dataset), Office-31, and Office-Home  \cite{saenko2010adapting,wang2018visual,rahman2019minimum}. These datasets are widely used in many publications \cite{gopalan2011domain,gong2012geodesic,wang2018visual}, and are the benchmarking data for evaluating the performance of domain adaptation algorithms. Table \ref{tab:data_des} lists the statistics of these datasets.
%($\#$ means the number). 
In the Office+Caltech-10 datasets and Office-31 dataset, there are totally four domains (A, W, C, and D) where A represents Amazon, W represents Webcam, C represents Caltech and D represents DSLR. In the Office-Home dataset, A represents Arts, C represents Clipart, P represents Product and R represents Real world.
C $\shortrightarrow$ A  means learning from existing domain C, and transferring knowledge to classify domain A. 

\begin{table}[h]
\small
\begin{center} 
\caption{Statistics of extracted IR features for four benchmark datasets}
 \setlength{\tabcolsep}{+0.5mm}{
\begin{tabular}{|c|c|c|c|c|c|c|c|c|c|c|c|}
\hline \label{tab:data_des}
Dataset & $\#$ Sample & $\#$ Feature & $\#$ Class & Domain(s) \\
\hline
Office-10  & 1410 & 1000 & 10 & A, W, D   \\
\hline
Caltech-10 & 1123 & 1000  & 10 & C  \\
\hline
Office-31 & 1330 & 1000 &31 & A, W, D   \\
\hline
Office-Home & 15588 & 1000 &65 & A, C, P, R  \\
\hline
\end{tabular}}
\end{center}
\end{table}

Fig.~\ref{fig:images} shows example images from three benchmark datasets. Amazon and Caltech images are mostly from online merchants, while DSLR  and Webcam images are mostly from offices \cite{gong2012geodesic}. We also combine Office-10 and Caltech-10 to be one dataset, and we perform twelve tasks in this dataset: C $\shortrightarrow$ A, C $\shortrightarrow$ W, $\cdots$, D $\shortrightarrow$ W. In Office-31 dataset, we have another six tasks: A $\shortrightarrow$ W, A $\shortrightarrow$ D, $\cdots$, D $\shortrightarrow$ W. For Office-Home datasets, we have another twelve tasks: A $\shortrightarrow$ C, A $\shortrightarrow$ P, $\cdots$, R $\shortrightarrow$ P.  Therefore, we have a total of 30 tasks in our experiment.

\begin{table}[h]
\small
\begin{center} 
\caption{Statistics of extracted IR features from different layers.}
 \setlength{\tabcolsep}{+0.3mm}{
\begin{tabular}{|c|c|c|c|c|c|c|c|c|c|c|c|}
\hline \label{tab:layers}
Layers &  Last average pooling& Last fully connected  & Classification \\
\hline
 $\#$ Feature  & 1536  & 1000  & 1000\\
\hline
\end{tabular}}
\end{center}
\end{table}

\begin{figure}[h]
\centering
\includegraphics[scale=0.46 ]{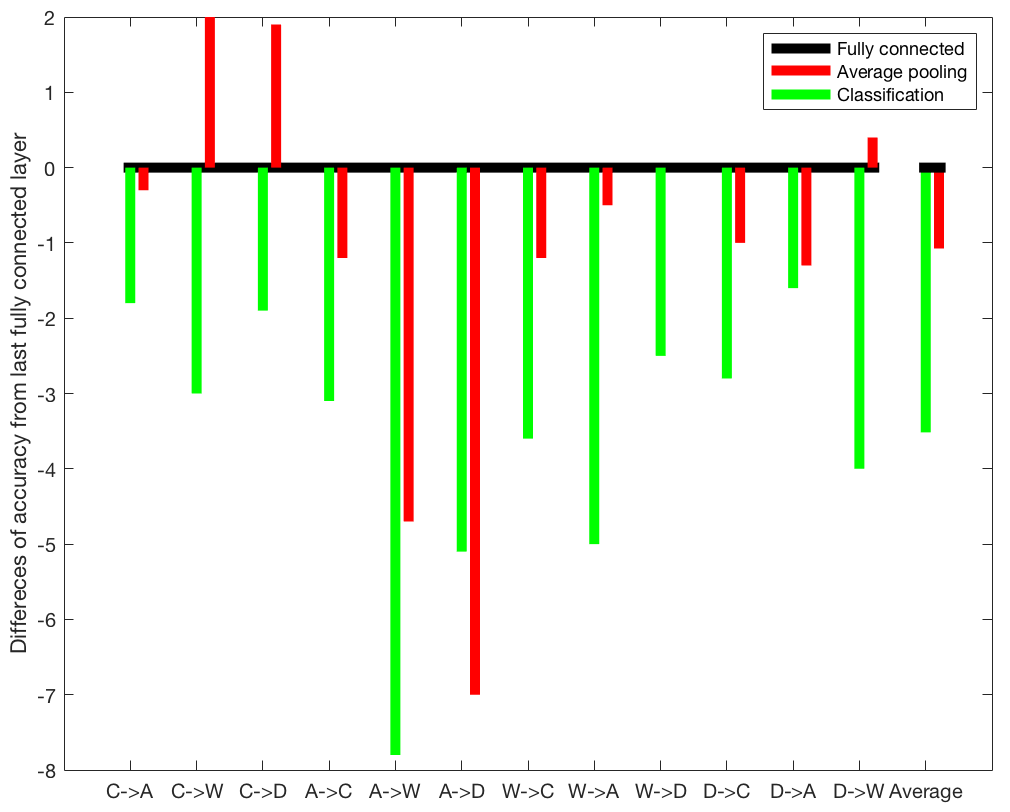}
\caption{Differences in accuracy of extracted features three layers (average pooling, fully connected, and classification) for the Office+Caltech-10 dataset tasks, where the baseline is the fully connected layer. The accuracy from the fully connected layer is better than other layers---all accuracies from the classification layer are below the fully connected layer, and most accuracies of the average pooling layer are below the fully connected layer. Therefore, we suggest extracting features from the last fully connected layer.}
\label{fig:layers}
\end{figure}

\begin{figure}[t]
\centering
\includegraphics[scale=0.5]{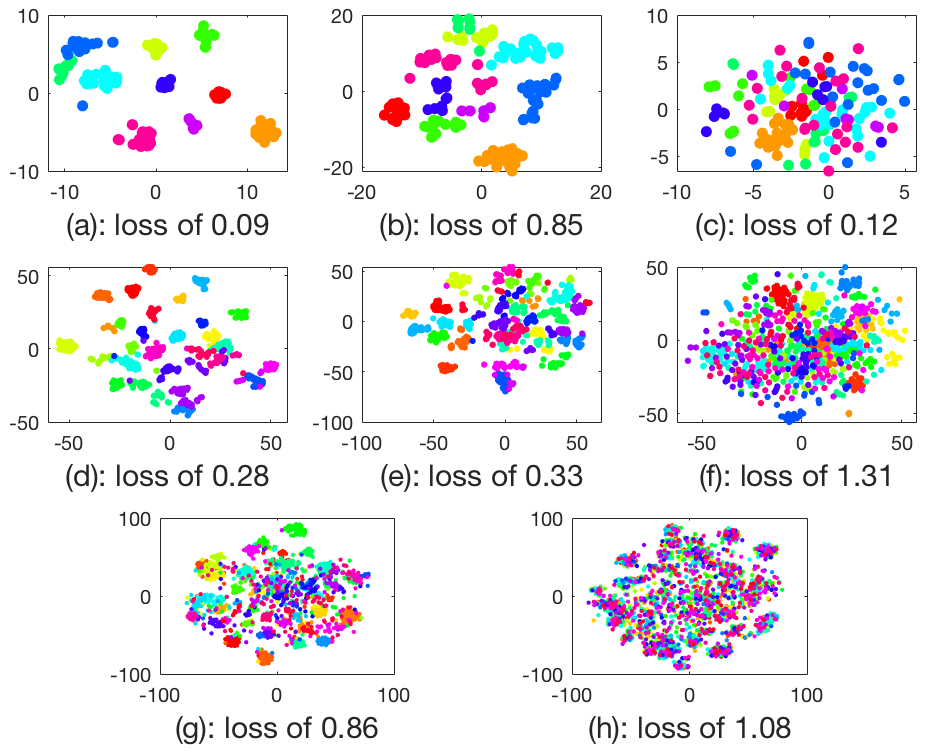}
\caption{The t-SNE view of the comparison of our IR features (a, d and g) with DeCAF (b and e), Resnet-50 (h), and SURF features (c and f).  Different color means different classes. The first row is from DSLR domain in Office+Caltech-10 datasets, and the second row is from the Webcam domain in the Office-31 dataset, and (g) and (h) are from the Art domain in Office-Home dataset. 
%And our IR feature has the lowest loss.
}
\label{fig:t-sne}
\end{figure}

\begin{table*}[h]
\begin{center}
\caption{Accuracy (\%) on Office + Caltech-10 datasets}
\setlength{\tabcolsep}{+1mm}{
\begin{tabular}{|c|c|c|c|c|c|c|c|c|c|c|c|c|c|c|c|c|}
\hline \label{tab:OC+10}
Task & C $\shortrightarrow$ A &  C $\shortrightarrow$ W & C $\shortrightarrow$ D & A $\shortrightarrow$ C & A $\shortrightarrow$ W & A $\shortrightarrow$ D & W $\shortrightarrow$ C & W $\shortrightarrow$ A & W $\shortrightarrow$ D & D $\shortrightarrow$ C & D $\shortrightarrow$ A & D $\shortrightarrow$ W & \textbf{Average}\\
\hline
TCA &		77&	80.7&	84.7	&	82.2	&68.1&	72.6&	79.3&	86.4&		88.5		&82.2	&	86.4&	84.7	&	81.1 \\
ITCA 	&	81	&65.8	&79.6	&	82.9&	70.8&	79&	78.2&	85.5&		92.4&		77.9	&	82.5&	90.5	&	80.5\\
SSTCA 	&	79.6&	70.5	&80.9&		76.5	&72.5&	83.4	&69.9&	79.5&		90.4&		78.7&		85.2&	87.8&		79.6\\
TJM 	&	86.7&	84.7&	86	&	82.8&	78.3&	86&	82	&86	&	\textbf{100}	&	83.8&		89.6&	99.3&		87.1\\
BDA &		89.5&	78.6&	81.5&		79.6&	73.2&	84.7&	78.1&	83.3&		\textbf{100}	&	79.7	&	88.5&	98.6&		84.6\\
JDA 	&	88.4&	84.4&	85.4&		81.6&	80.7&	81.5&	82.2&	89.8&		\textbf{100}	&	86	&	91.5&	99.3&		87.6\\
SVM 	&	91	&78 &	85.4	&	83.3&	72.5&	83.4	&62.9&	72.1	&	99.4&		65	&	78.2&	96.6	&	80.7\\
GFK 	&	88.8&	77.3&	86	&	77.4&	66.8&	79&	72	&76.5&		\textbf{100}	&	75.5&		84.7	&99	&	81.9\\
JGSA & 91.4 & 86.8 & 93.6 & 84.9 & 81.0 & 88.5 & 85.0 & 90.7 & \textbf{100} & 86.2 & 92.0 & 99.7 & 90.0\\
ARTL & 92.4 & 87.8 & 86.6 & 87.4 & 88.5 & 85.4 & 88.2 & 92.3 & \textbf{100} & 87.3 & 92.7 & \textbf{100} & 90.7\\
MEDA 	&	93	&91.2&	89.8&		89	&90.8	&88.5&	89&	92.2	&	99.4&		88.6&		93.2&	98.6&		91.9\\
AlexNet 	&	91.9&	83.7	&87.1&		83	&79.5&	87.4&	73	&83.8	&	\textbf{100}&	79	&	87.1&	97.7&		86.1\\
DAN 	&	92&	90.6&	89.3	&	84.1&	91.8&	91.7&	81.2&	92.1&		\textbf{100}	&	80.3	&	90&	98.5&		90.1\\
DDC 	&	91.9&	85.4&	88.8&		85&	86.1&	89&	78&	83.8&		\textbf{100}	&	79	&	87.1&	97.7	&	86.1\\
DCORAL 	&	89.8&	\textbf{97.3} &	91	&	91.9&	\textbf{100}	&90.5&	83.7&	81.5&		90.1	&88.6	&	80.1&	92.3&		89.7\\
 \hline
 \hline
 \textbf{MEDA-IR}	& \textbf{96.2}	&95.9&	96.2	& \textbf{95.2} &	98	&96.8&	94.5&	96.2&	99.4&	93.8&	95.5&	98.6 & 96.4\\
\textbf{MDAIR}	&	96.1 &	94.9&	\textbf{96.2} &		94.2 &	98.6 &	\textbf{100}	&\textbf{94.9}  & \textbf{96.3} &		\textbf{100}	&	\textbf{94.2} &		\textbf{95.8} &	98.6&		\textbf{96.7}	\\
\hline
\end{tabular}}
\end{center}
\end{table*}

\begin{table*}[ht]
\begin{center}
\caption{Accuracy (\%) on Office-31 datasets}
\setlength{\tabcolsep}{+1.8mm}{
\begin{tabular}{|c|c|c|c|c|c|c|c|c|c|c||c|c|}	
\hline \label{tab:O31}
Task & TCA & SSTCA & MEDA & DAN  & RTN & DANN & ADDA & CAN  &JDDA  & JAN &\textbf{MEDA-IR} & \textbf{MDAIR}\\
\hline
A $\shortrightarrow$ W  &82.6 & 81 &83.3  & 80.5& 84.5& 82 & 86.2	&81.5& 82.6 & 85.4 & 90.8 &\textbf{94}\\
\hline
A $\shortrightarrow$ D &84.1 &78.7 & 83.3 & 78.6&77.5 &79.7 & 77.8 &65.9 &79.8 & 84.7 & 91.4 &\textbf{92.6}\\
\hline
W $\shortrightarrow$ A &69.1 &68.9 & 66.2 &62.8 &64.8 &67.4 &68.9 &\textbf{98.2} &66.7 &70.0 & 74.6 &77.6\\
\hline
W $\shortrightarrow$ D &99.6 &99.6 & 96  & 99.6 &99.4 &99.1 &98.4 &85.5 & 99.7 &\textbf{99.8} & 97.2 &99.2\\
\hline
D $\shortrightarrow$ A &66.1 &66.6 &66.7  &63.6 &66.2 &68.2 &69.5 & \textbf{99.7} &57.4 &68.6 & 75.4 &78.7\\
\hline
D $\shortrightarrow$ W &97 & \textbf{97.4} & 91.7  &97.1 &96.8 &96.9 &96.2 &63.4 &95.2  &\textbf{97.4} & 96 &96.9\\
\hline
\hline
\textbf{Average}  & 83.1&82.0 &81.2  &80.4 &81.6 &82.2 &82.9 &82.4 &80.2 &84.3 & 87.5 &\textbf{89.8}\\
\hline
\end{tabular}}
\end{center}
\end{table*}

\begin{table*}[h]
\begin{center}
\caption{Accuracy (\%) on Office-Home datasets}
\setlength{\tabcolsep}{+1.2mm}{
\begin{tabular}{|c|c|c|c|c|c|c|c|c|c|c|c|c||c|}
\hline \label{tab:OH}
\bf Task & A $\shortrightarrow$ C &  A $\shortrightarrow$ P & A $\shortrightarrow$ R & C $\shortrightarrow$ A & C $\shortrightarrow$ P & C $\shortrightarrow$ R & P $\shortrightarrow$ A & P $\shortrightarrow$ C & P $\shortrightarrow$ R & R $\shortrightarrow$ A & R $\shortrightarrow$ C & R $\shortrightarrow$ P & \textbf{Average}\\
\hline
AlexNet	& 26.4& 	32.6& 	41.3& 	22.1& 	41.7& 	42.1& 	20.5& 	20.3& 	51.1& 	31& 	27.9	& 54.9& 	34.3 \\
VGG16	& 30.4 &45.9 & 57.5  &35.4 & 48.7 &50.8 &35.8 &30.5 &60.2 &49.6 & 34.5 & 64.0 &45.3 \\
D-CORAL & 32.2 &40.5 &54.5 & 31.5 & 45.8 &47.3 &30.0 &32.3 &55.3 & 44.7 & 42.8 &59.4 &42.8 \\
RTN  &31.3 &40.2  & 54.6 &32.5 &46.6 &48.3 &28.2 &32.9 &56.4 &45.5 &44.8 &61.3 &43.5\\
DAH & 31.6 &40.8 &51.7 &34.7 &51.9 &52.8 &29.9 &39.6 &60.7 &45.0 &45.1 &62.5 &45.5\\
MDDA& 35.2 &44.4 &57.2 &36.8 & 52.5 &53.7 &34.8 &37.2 &62.2 &50.0 &46.3 &66.1 &48.0\\
% DAN	& 31.7& 	43.2& 	55.1& 	33.8& 	48.6& 	50.8& 	30.1& 	35.1& 	57.7& 	44.6& 	39.3& 	63.7& 	44.5\\
% DANN	& 36.4	& 45.2& 	54.7& 	35.2& 	51.8& 	55.1& 	31.6& 	39.7& 	59.3& 	45.7& 	46.4& 	65.9& 	47.3\\
% JAN	& 35.5	& 46.1& 	57.7& 	36.4& 	53.3	& 54.5& 	33.4& 	40.3& 	60.1& 	45.9& 	47.4& 	67.9& 	48.2\\
% CDAN-RM	& 36.2& 	47.3& 	58.6& 	37.3& 	54.4& 	58.3 	& 33.2	& 43.9& 	62.1& 	48.2& 	48.1	& 70.7	& 49.9\\
% CDAN-M	& 38.1	& 50.3& 	60.3& 	39.7& 	56.4& 	57.8& 	35.5& 	43.1& 	63.2& 	48.4& 	48.5& 	71.1& 	51\\
ResNet-50 & 	34.9& 	50	& 58& 	37.4& 	41.9& 	46.2& 	38.5& 	31.2& 	60.4& 	53.9& 	41.2& 	59.9& 	46.1\\
DAN	& 43.6	& 57& 	67.9& 	45.8& 	56.5& 	60.4& 	44& 	43.6& 	67.7& 	63.1& 	51.5& 	74.3& 	56.3\\
DANN	& 45.6	& 59.3& 	70.1& 	47& 	58.5& 	60.9& 	46.1& 	43.7& 	68.5& 	63.2& 	51.8& 	76.8& 	57.6\\
JAN	& 45.9& 	61.2& 	68.9& 	50.4& 	59.7& 	61& 	45.8& 	43.4& 	70.3& 	63.9& 	52.4& 	76.8& 	58.3\\
CDAN-RM	& 49.2& 	64.8& 	72.9& 	53.8& 	62.4& 	62.9& 	49.8& 	48.8& 	71.5& 	65.8& 	56.4& 	79.2& 	61.5\\
CDAN-M	& 50.6& 	65.9& 	73.4& 	55.7& 	62.7& 	64.2& 	51.8& 	49.1& 	74.5& 	68.2& 	56.9& 	80.7& 	62.8\\
\hline
\hline
\textbf{MEDA-IR}	& 52.9&	79.3&	78.9&	67.3&	78.8&	78.8&	68.2&	53.4&	79.8&	71.8&	56.3&	83& 	70.7\\
% \hline
\textbf{MDAIR} & \textbf{55.6}&	\textbf{80.4}&	\textbf{81.6}&	\textbf{70.2}&	\textbf{80.7}&	\textbf{80.8}&	\textbf{71}&	\textbf{55.6}&	\textbf{82.5}&	\textbf{73.5}&	\textbf{57.7}	& \textbf{83.9} & \textbf{72.8}\\
\hline
\end{tabular}}
\end{center}
\end{table*}

\subsection{Feature Comparison} \label{sec:fea_c}

To determine the best layer for feature extraction, we first explore the effect of different layers in final accuracy.  We list the number of features from three layers in Tab.~\ref{tab:layers}. Based on an experiment using the Office+Caltech-10 dataset, we choose the optimal layer. Fig.~\ref{fig:layers} shows the accuracy of different tasks from different layers. Accuracy from the fully connected layer is typically higher than the other two layers. Therefore, we suggest that last fully connected layer is the best layer to extract features in domain adaption problem.

We then examine the quality of our IR features in the last fully connected layer. 
%We first visualize the histogram of our IR data. As shown in Fig.~\ref{fig:hist},  it compares the histogram of our IR feature with SURF and DeCAF feature. It is clear that the IR feature is well separated than the other two features. Secondly, w
We visualize the three domains from three datasets using the t-SNE technique. T-SNE (t-distributed Stochastic Neighbor Embedding) \cite{maaten2008visualizing} is an algorithm for visualizing high-dimensional data by re-representing it in a lower dimensional space. 
t-SNE generates a low-dimensional representation in which points near each other are similar in the high-dimensional space and vice versa.
%The similarities between points in low dimensions are intended to be similar to that of the original high dimensional representation. In addition, the nearby points in the low-dimensional space correspond to nearby points in high-dimensional space. 
The better that clusters are separated in the t-SNE view, the better the extracted features are likely to be. 
% and distant points in high-dimensional space correspond to distant embedded low-dimensional points. 
The loss function of the t-SNE method is  Kullback-Leibler divergence, which measures the difference between the two distributions \cite{maaten2008visualizing}. 
%We have found that
Typically, lower losses correspond to better features.
%Also, the loss is an indicator to illustrate the goodness of the feature (the lower loss means the better of features).

As shown in Fig.~\ref{fig:t-sne}, our extracted IR features produce better clearer and better separated clusters
%show a well separated distribution 
than SURF, DeCAF, and Resnet-50 features. Therefore, we can assume that our IR features will lead to better classification result than the others. Similarly, the visualizations based on our IR features have the lowest loss values.

\subsection{Comparison to State-of-the-art Methods}

We compare the performance of our MDAIR model with 25 state-of-the-art (both traditional and deep learning) methods:
Transfer Component Analysis  (TCA) \cite{pan2011domain}; 
Global  and  Local  Metrics  for  Domain  Adaptation (IGLDA also called ITCA)  \cite{jiang2017integration};  
Semi-supervised TCA (SSTCA)  \cite{pan2011domain};  
Transfer Joint Matching (TJM) \cite{long2014transfer};  
Balanced distribution adaptation (BDA) \cite{wang2017balanced}; 
Joint distribution alignment (JDA) \cite{long2013transfer}; 
Support Vector Machine (SVM) \cite{bergamo2010exploiting}; 
Geodesic Flow Kernel (GFK) \cite{gong2012geodesic};
Adaptation Regularization (ARTL) \cite{long2014adaptation};
Joint Geometrical and Statistical Alignment (JGSA)
\cite{zhang2017joint}; 
Manifold Embedded Distribution Alignment (MEDA) \cite{wang2018visual}; 
AlexNet \cite{krizhevsky2012imagenet}; 
VGG-16 \cite{simonyan2014very};
Deep Adaptation Networks (DAN) \cite{long2015learning}; 
Deep Domain Confusion (DDC) \cite{tzeng2014deep}; Deep  Correlation Alignment 
(DCORAL) \cite{sun2016deep}; 
Joint Adaptation Networks (JAN) \cite{long2017deep};
Residual Transfer Networks (RTN) \cite{long2016unsupervised};
Domain Adaptive Neural Networks (DANN) \cite{ghifary2014domain};
Domain Adaptive Hashing
(DAH) \cite{venkateswara2017deep};
Minimum Discrepancy Deep Adaptation (MDDA) \cite{rahman2019minimum};
Adversarial Discriminative Domain Adaptation (ADDA) \cite{tzeng2017adversarial};
Collaborative Adversarial Network (CAN) \cite{zhang2018collaborative},  Joint Discriminative  Domain Adaptation (JDDA) \cite{chen2018joint}, and Conditional Domain Adversarial Networks (CDAN-RM, CDAN-M) \cite{long2018conditional}.

 From Tables~\ref{tab:OC+10}, \ref{tab:O31} and \ref{tab:OH}, we can observe that the accuracy of MDAIR model is ahead of all other methods in most tasks (23/30).  Notably, our model always achieves the best performance in Office-Home dataset.  Regarding all three datasets, the overall average performance is significantly improved over the best state-of-the-art baseline methods. The results of using SURF feature are too low to compare with DeCAF and IR features and are omitted.
 %, we do not list the results in our comparison. 

\begin{figure}[t]
\centering
\includegraphics[scale=0.5]{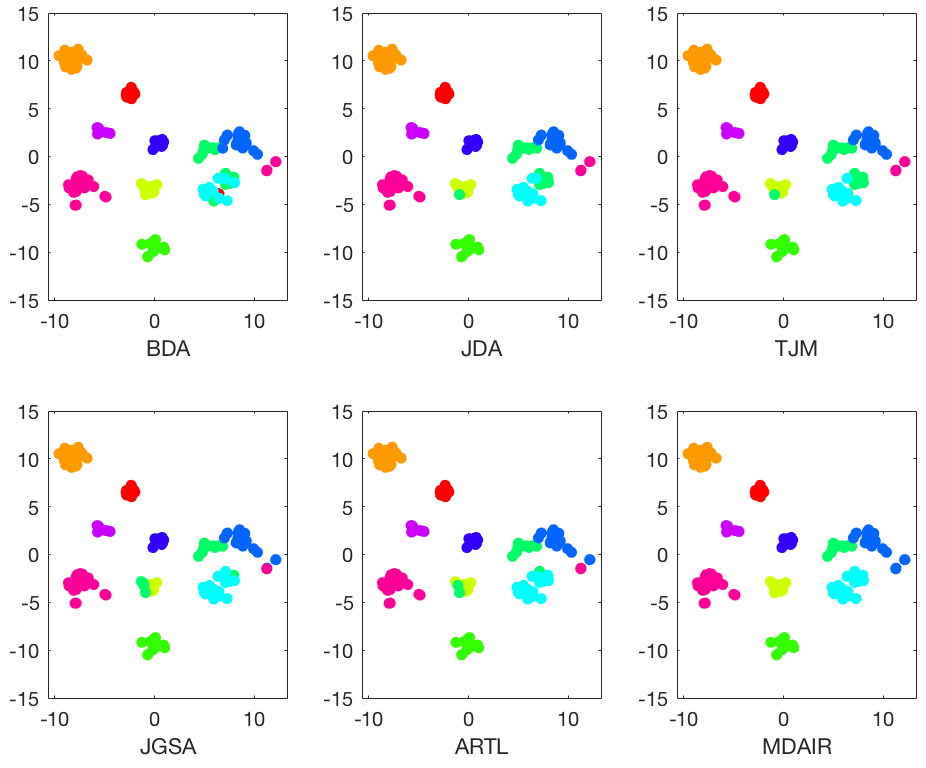}
\caption{T-SNE view of the comparison of baseline methods and the proposed MDAIR model in the A $\shortrightarrow$ D in Office+Caltech-10 dataset. The proposed MDAIR model has the highest accuracy, while all other methods have some mixed colors, which implied the classes are wrongly classified (as colors correspond to labels).}
\label{fig:all_methods}
\end{figure}

To illustrate the effectiveness of our model, we consider the case in which all models use our IR features, and view the prediction results using t-SNE. Focusing on the A$\shortrightarrow$D task in which the accuracy of our MDAIR is 100\% (and thus identical to ground truth), 
Fig.~\ref{fig:all_methods} shows that all other conventional methods contained  mixed colors in the t-SNE view. These results indicate that our modified distribution alignment is better than several baseline methods even using the same features. In addition, we test our IR features using the original MEDA method (MEDA-IR in Tables~\ref{tab:OC+10}, \ref{tab:O31} and \ref{tab:OH}); results still turn out that our modified distribution alignment is better than the previous MEDA model.

\subsection{Parameter Settings }
In our experiments, the optimal parameters for different tasks might be different. To more easily reproduce our results, we use consistent parameters: $\lambda=10$, $\rho=1.0$, $p=10$, and $\eta=0.1$.

\section{Discussion}

\begin{table}[h]
\begin{center}
\caption{Comparison of average accuracy of the best baseline method and our MDAIR model}
 \setlength{\tabcolsep}{+0.7mm}{
\begin{tabular}{|c|c|c|c|} 
\hline \label{tab:space}
Task & Best baseline & MDAIR & Improvement \\
\hline
Office+Caltech-10   & 91.9 & 96.7 & 4.8\%   \\
\hline
Office-31  & 84.3 & 89.8  & 5.5\%   \\
\hline
Office-Home   & 62.8 & 72.8 &10\%  \\
\hline
\end{tabular}}
\end{center}
\end{table}

We list the improvement of our model based on the best state-of-the-art methods in Table~\ref{tab:space}. For three datasets (Office+Caltech-10, Office-31, and Office-Home), our method improves the absolute accuracy by 4.8\%, 5.5\%, and 10\% respectively. Therefore, the quality of our model exceeds that of all the state-of-the-art methods.

There are two prominent reasons for the success of our model. First of all, our model takes advantage of deep features from the Inception-ResNet-v2 model, which produces  better features than SURF and DeCAF features. And better features reduce the difference between the source and target domains. Secondly, the modified distribution alignment facilitate the alignment of the distribution of features which leads to higher accuracy. 

In addition, our experiments imply that the last fully connected layer is the best layer for feature extraction. A likely reason is that the layer collects all features from the previous layer; hence it will form better features than previous layer. Although the last classification layer can be used for feature extraction from the IR model, the performance is worse than the last fully connected layer since features from classification layer will be affected by original trained classes. We observe that our model is compromised in some tasks (A $\shortrightarrow$ W in Office+Caltech-10 and D $\shortrightarrow$ A in Office-31 dataset). This caused by the intrinsic differences of  various datasets, and so we cannot guarantee that our model always beats all other methods.

However, one shallow weakness of our model is that feature extraction affects the results significantly. We suggest that extracting feature from higher top-1 accuracy deep neural networks will further improve the accuracy.

\section{Conclusion}
In this paper, we are the first to extract features from a pre-trained Inception-ResNet-v2 model for the domain adaption problem.  The experiment shows that the last fully connected layer is the best layer to extract features and the extracted features are better than DeCAF and Resnet-50 features.  The modified distribution alignment model has a better performance than other models. We also test our model using three benchmark datasets.   Extensive experiments demonstrate significant improvements in classification accuracy over the state-of-the-art. 

There are some obvious areas for follow-up work. Extracting features from another well-trained deep neural network might generate a better input for the modified distribution alignment than the IR model.  Testing on a broader set of unsupervised learning tasks will improve the applicability of our model. Also, a new distribution method will be beneficial for increasing the predictive accuracy.

%-------------------------------------------------------------------------

\footnotesize
\bibliographystyle{unsrt}
\bibliography{references}

% that's all folks
\end{document}